\newif\ifarxiv
\arxivtrue

\documentclass[final,12pt]{clear2025} %

\title[Causal Responsibility Attribution for Human-AI Collaboration]{Causal Responsibility Attribution for Human-AI Collaboration
}

\usepackage{times}
\numberwithin{equation}{section}

\usepackage{fancyhdr}

\fancypagestyle{firstpagefooter}{
    \fancyhead[]{} %
    \fancyfoot[L]{Preprint. Under review.} %
}

\clearauthor{%
 \Name{Yahang Qi} \Email{yahaqi@ethz.ch}\\
 \addr ETH Z\"urich
 \AND
 \Name{Bernhard Sch\"olkopf} \Email{bs@tuebingen.mpg.de} \\
 \addr MPI T\"ubingen
 \AND
 \Name{Zhijing Jin} \Email{zjin@cs.toronto.edu}\\
 \addr University of Toronto%
}

\begin{document}

\maketitle

\begin{abstract}%
    As Artificial Intelligence (AI) systems increasingly influence decision-making across various fields, the need to attribute responsibility for undesirable outcomes has become essential, though complicated by the complex interplay between humans and AI. Existing attribution methods based on actual causality and Shapley values tend to disproportionately blame agents who contribute more to an outcome and rely on real-world measures of blameworthiness that may misalign with responsible AI standards. This paper presents a causal framework using Structural Causal Models (SCMs) to systematically attribute responsibility in human-AI systems, measuring overall blameworthiness while employing counterfactual reasoning to account for agents' expected epistemic levels. Two case studies illustrate the framework’s adaptability in diverse human-AI collaboration scenarios.\footnote{Our code and data 
\ifarxiv
are at \url{https://github.com/yahang-qi/Causal-Attr-Human-AI.git}.
\else
have been uploaded to the submission system, and will be open-sourced upon acceptance.
\fi
}
\end{abstract}


\thispagestyle{firstpagefooter} 

\section{Introduction}

As Artificial Intelligence (AI) systems increasingly influence decision-making in critical sectors such as healthcare \citep{budd2021survey}, finance \citep{cohen2023arbitrage}, and autonomous driving \citep{badue2021self}, the need to clearly define and attribute responsibility when outcomes are undesirable or failures occur becomes crucial \citep{santoni2018meaningful}. The integration of AI complicates traditional accountability mechanisms due to the shared decision-making responsibilities between humans and algorithms. Human-AI systems pose unique challenges for responsibility attribution, given their interactive, complex, and high-stakes nature \citep{amershi2019guidelines}. On one hand, the ability of humans to override or modify AI-driven decisions introduces an element of unpredictability in outcomes. On the other hand, AI decisions often rely on huge datasets or complex models that lack full transparency, making it difficult for human counterparts to fully understand or anticipate AI behaviour.

To address these challenges, various methods have been proposed for responsibility attribution in multi-agent settings. \citet{chockler2004responsibility} and \citet{halpern2018towards} introduced frameworks to quantify blameworthiness based on causal relationships. This definition was further extended to multi-agent settings by attributing blameworthiness through the Shapley value, as demonstrated by \citet{friedenberg2019blameworthiness}. Additionally, responsibility attribution has been studied in decentralized, partially observable Markov decision processes using the concept of actual causality \citep{triantafyllou2022actual}. These approaches rely on actual causality \citep{halpern2016actual} to measure the degree of blameworthiness among agents.

However, in human-AI collaboration, using actual causality and the Shapley value presents limitations. One issue is that agents who contribute more to a task are assigned a higher degree of blame, which may not align with intuitive notions of responsibility in collaborative settings \citep{kumar2020problems}. Furthermore, these frameworks define blameworthiness based on the real-world probability measure. For trustworthy and reliable AI, it is essential to measure blameworthiness against the epistemic standards that AI systems should uphold, rather than solely on actual causal effects \citep{bostrom2018ethics}.

This paper presents a causal framework based on Structural Causal Models \citep[SCM;][]{pearl2009causality} for systematically attributing responsibility in human-AI systems. The framework is designed to be adaptable across diverse human-AI collaboration scenarios, ensuring broad applicability and relevance. First, the paper defines a measure of blameworthiness for different actions and their associated outcomes, building on the foundations established in Halpern's framework \citep{halpern2018towards}. Using counterfactual reasoning, the framework then attributes responsibility for specific undesirable outcomes to the relevant parties, taking into account the epistemic level each party should possess. Finally, two case studies are provided to demonstrate the practical application of this responsibility attribution framework.

\section{Related Work}

\subsection{Responsibility Attribution}

The need for judging moral responsibility arises both in ethics and in law. However, these notions are notoriously difficult to define carefully. One famous example is the trolley-problem \citep{thomson1984trolley}, which is a moral dilemma in which one must decide whether to pull a lever to divert a runaway trolley onto a track where it will kill one person, rather than allowing it to continue on its current track, where it will kill five. To formally define moral responsibility, there is general agreement that a definition of moral responsibility will require integrating causality \citep{chockler2004responsibility}, intention \citep{cushman2015deconstructing}, knowledge \citep{malle2014theory}. To formally define these, structural causal models provide a powerful tool \citep{halpern2005causes}. Using this framework, \citet{halpern2018towards} defined the degree of blameworthiness in terms of actual causality. Furthermore, the degree of blameworthiness can be discounted by the cost of the action. This notation of blameworthiness is further generalised into multi-agent setting by ascribing blameworthiness to the group of agents relative to an epistemic state of potential outcomes \citep{friedenberg2019blameworthiness}. The degree of blameworthiness for individuals can be attributed using the Shapley value, a notion from cooperative game theory \citep{shapley1953value}. However, when using Shapley value for responsibility attribution, agents which contributes more to the task may end up being blamed more, which may not align with intuitive notions of responsibility in collaborative settings \citep{kumar2020problems}.  

In the regime of AI, \cite{franklin2022causal} outlined a causal framework of responsibility attribution which integrates nine factors: causality, role, knowledge, objective foreseeability, capability, intent, desire, autonomy, and character. The causal relationship between these nice factors and the responsibility is extensively discussed. However, no explicit mathematical formulation is given. Regarding multi-agent setting, \citet{triantafyllou2022actual} proposed a way to attribute the responsibility in decentralized partially observable Markov decision processes. However, these frameworks do not account for agents’ expected epistemic levels, which is not helpful to build reliable and trustworthy AI systems \citep{bostrom2018ethics}.

\subsection{Human-AI Collaboration}

The integration of AI in critical fields like healthcare \citep{budd2021survey}, finance \citep{cohen2023arbitrage}, and autonomous driving \citep{badue2021self} has driven the development of structured human-AI collaboration frameworks \citep{wang2020human}. These frameworks generally fit into three modes, which define the balance of human and AI roles: AI-centric, Human-centric, and Symbiotic \citep{fragiadakis2024evaluating}.

In the AI-centric mode, AI systems take the lead, performing tasks with minimal human input. This mode focuses on maximizing computational efficiency, where AI operates autonomously in contexts like agentic systems \citep{shavit2023practices} or complex data predictions, such as protein structures \citep{jumper2021highly}. Human-centric frameworks, often called human-in-the-loop, maintain human oversight as central, employing AI as an auxiliary tool to manage data-heavy or repetitive tasks. This mode supports fields where human judgment is crucial, like diagnostics in healthcare \citep{chaddad2023survey} and assisted driving \citep{badue2021self}. The Symbiotic mode promotes a balanced partnership, with human and AI systems sharing decision-making and complementing each other’s strengths. This setup enables mutual feedback and close collaboration, ideal for tasks requiring both AI’s computational power and human intuition, such as creative co-production \citep{rezwana2023designing}.

As these systems become more prevalent, they underscore the need for responsibility attribution frameworks that address the complexity of human-AI interactions and evaluate responsibility against the epistemic standards expected from trustworthy AI systems.epistemic standards necessary for reliable and trustworthy AI systems..

\section{Preliminaries}
This section introduces the fundamental concepts and mathematical tools utilised in our study, specifically focusing on Structural Causal Models (SCMs), interventional counterfactual, and backtracking counterfactual.

\subsection{Structural Causal Models (SCMs)}

SCMs \citep{pearl2009causality} offer a structured framework for modelling and analysing causal relationships between variables. We introduce SCMs below, using notation adapted from \cite{bongers2021foundations} and \cite{peters2017elements}.

\begin{definition}[Structural Causal Model \citep{bongers2021foundations}]
    A structural causal model (SCM) is a tuple $\mathcal{M} = \langle \mathcal{I}, \mathcal{J}, \mathcal{X}, \mathcal{E}, f, \mathbb{P}(\mathcal{E})\rangle$, where
        \begin{itemize}
            \item $\mathcal{I}$ and $\mathcal{J}$ are disjoint finite index set of endogenous and exogenous variables, respectively.
            \item The domains $\mathcal{X} = \prod_{i\in\mathcal{I}}\mathcal{X}_i$ and $\mathcal{E} = \prod_{j\in\mathcal{J}}\mathcal{E}_i$ are products of standard Borel space.
            \item The exogenous distribution $\mathbb{P}(\mathcal{E}) = \bigotimes_{j\in\mathcal{J}}\mathbb{P}(\mathcal{E}_j)$ is the product of probability distributions.
            \item The causal mechanism $f:\mathcal{X}\times\mathcal{E}\to\mathcal{X}$ is a measurable function.
        \end{itemize}
\end{definition}
The solutions of an SCM in terms of random variables are defined up to almost sure equality.

\begin{definition}[Solution to an SCM \citep{bongers2021foundations}]
    A pair $(\mathbf{X}, \mathbf{E})$ of random variables $\mathbf{X}: \Omega\to\mathcal{X}$, $E: \Omega\to\mathcal{E}$, where $\Omega$ is a probability space, is a solution of the SCM $\mathcal{M} = \langle \mathcal{I}, \mathcal{J}, \mathcal{X}, \mathcal{E}, f, \mathbb{P}(\mathcal{E})\rangle$ if
        \begin{itemize}
            \item $\mathbb{P}^\mathbf{E} = \mathbb{P}_\mathcal{E}$, that is, the distribution of $\mathbf{E}$ is equal to $\mathbb{P}_\mathcal{E}$, and
            \item the structural equations are satisfied, that is,
        \begin{align}
            \mathbf{X} = f(\mathbf{X}, \mathbf{E})\ a.s..
        \end{align}
        \end{itemize}
    For convenience, we call a random variable $\mathbf{X}$ a solution of $\mathcal{M}$ if there exists a random variable $\mathbf{E}$ such that $(\mathbf{X}, \mathbf{E})$ forms a solution of $\mathcal{M}$.
\end{definition}
In this paper, we restrict our attention to acyclic SCMs, where there is a total ordering of the endogenous variables. In this setting, given a setting $\mathbf{e}$ for the exogenous variables, there is a unique solution $\mathbf{x}_\mathbf{e}$ for all the structural equations.

\subsection{Counterfactual}

Counterfactual is used to reason about what would happen to a system if we were to intervene and change some of its variables \citep{peters2017elements}. In the context of an SCM, an intervention that sets a variable $X$ to $x$ is denoted as $do(X = x)$. The effect of this intervention on another variable $Y$ is computed by modifying the structural equations in the SCM and propagating the effects.

\begin{definition}[Counterfactuals]
Consider an SCM $\mathcal{M} = \langle \mathcal{I}, \mathcal{J}, \mathcal{X}, \mathcal{E}, f, \mathbb{P}(\mathcal{E})\rangle$ and its solution $\mathbf{X}$. Given some observations $\mathbf{x}\in\mathcal{X}$, we define a counterfactual SCM $\mathcal{M}' = \langle \mathcal{I}, \mathcal{J}, \mathcal{X}, \mathcal{E}, f, \mathbb{P}(\mathcal{E}|\mathbf{X}=\mathbf{x})\rangle$ by replacing the distribution of noise variables with the distribution conditioned on observation $\mathbf{X}=\mathbf{x}$, denoted as $\mathbb{P}(\mathcal{E}|\mathbf{X}=\mathbf{x})$.
\end{definition}

\section{A Causal Framework of Responsibility Attribution }\label{sec:framework}

This section presents the formal definitions and mathematical formulations for undesirable outcomes and the degree of blameworthiness of an action relative to an alternative action, both of which are fundamental to our causal responsibility attribution framework. We begin by defining these key terms, followed by a formal definition of the overall degree of blameworthiness in a human-AI decision-making system. Finally, we focus on specific outcomes and demonstrate how to attribute blameworthiness to different parties based on their epistemic levels, aligned with responsible AI standards and represented by a probability measure over potential outcomes.. 

\subsection{Degree of Blameworthiness}
Consider a system whose causal mechanism is defined by the SCM $\mathcal{M} = \langle \mathcal{I}, \mathcal{J}, \mathcal{X}, \mathcal{E}, f, \mathbb{P}(\mathcal{E})\rangle$. In this system, an agent 
takes action and may affect the probability of occurrence of some unwanted outcome $\phi$, which is defined as
\begin{definition}[Outcome in an SCM] Consider an SCM $\mathcal{M} = \langle \mathcal{I}, \mathcal{J}, \mathcal{X}, \mathcal{E}, f, \mathbb{P}(\mathcal{E}) \rangle$ with a solution $\mathbf{X}$. An outcome $\phi$ is defined as a measurable function $\phi: \mathcal{X} \to \{0, 1\}$, where $\phi = 1$ if and only if $\mathbf{X}$ takes a specified value or falls within a designated subset of $\mathcal{X}$. That is, the outcome $\phi = 1$ "happens" when $\mathbf{X}$ satisfies certain conditions and $\phi = 0$ otherwise. Its probability is then defined in the following:
\begin{align}
\mathbb{P}(\phi = 1) = \mathbb{P}\left(\left\{\mathbf{e}\in\mathcal{E}: \phi(\mathbf{x}_\mathbf{e} = 1)\right\}\right),
\end{align}
where $\mathbf{x}_\mathbf{e}$ is the solution under causal setting $\mathbf{e}$.
\end{definition}

\begin{figure}
    \centering
    \includegraphics[width=0.3\linewidth]{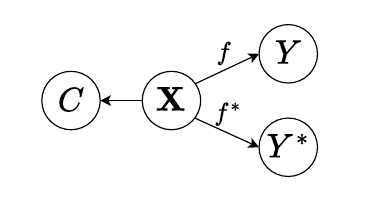}
    \caption{SCM of an Decision-Making System.}
    \label{fig:Decisin_Making_System}
\end{figure}
In decision-making systems, an unwanted outcome could be failure in making correct decision. As shown in Figure \ref{fig:Decisin_Making_System}, a decision $Y$ made from available information $\mathbf{X}$, with $f$ being decision-making policy. 
The optimal decision $Y^*$ is made using optimal policy $f^*$. 
An action $a$ modifies the decision-making system $\mathcal{M}$ by modifying its decision-making policy, denoted as $f^a$. Such action may or may not increase the probability of unwanted outcome $\phi$, when compared with another action $a'$.
\begin{definition}[Blameworthiness] Consider a decision-making system represented by  SCM $\mathcal{M} = \langle \mathcal{I}, \mathcal{J}, \mathcal{X}, \mathcal{E}, f, \mathbb{P}(\mathcal{E}) \rangle$ with a solution $\mathbf{X}$, consider action $a$ and reference action $a'$, action $a$ is said to be blameworthy for outcome $\phi$ when compared with the reference action $a'$ when it increases the probability of $\phi$:
\begin{align}
\delta(a, a') = \mathrm{max}\left\{0,\ \mathbb{P}(\phi=1|\mathcal{M}^a) - \mathbb{P}(\phi=1|\mathcal{M}^{a'})\right\}
~, 
\end{align}
where $\mathcal{M}^a = \langle \mathcal{I}, \mathcal{J}, \mathcal{X}, \mathcal{E}, f^a, \mathbb{P}(\mathcal{E})\rangle$ and $\mathcal{M}^{a'} = \langle \mathcal{I}, \mathcal{J}, \mathcal{X}, \mathcal{E}, f^{a'}, \mathbb{P}(\mathcal{E})\rangle$ are decision-making system modified via action $a$ and $a'$, respectively.  
\end{definition}
We can determine the extent to which agent performing action $a$ affects the probability of outcome $\phi$. 
This effect is compared with another possible action $a'$, which is different from $a$. 
However, in real-world scenarios, a decision-making system is often modified by an action $a$ because this can reduce the cost or improve the efficiency of the system. 
We use random variable $C$ to measure the cost of a decision $C: \mathcal{X}\to[0, \infty)$. 
As a result, the degree of blameworthiness is discounted by the improvement in efficiency.

\begin{definition}[Discounted Blameworthiness] 
Consider a decision-making system represented by SCM $\mathcal{M} = \langle \mathcal{I}, \mathcal{J}, \mathcal{X}, \mathcal{E}, f, \mathbb{P}(\mathcal{E}) \rangle$ modified by action $a$ and $a'$. 
The discounted degree of blameworthiness is defined by
\begin{align}
\mathrm{DB}(a, a') = \gamma(a, a')\delta(a, a')
~, 
\end{align}
where $\gamma(a, a') = \tau(\mathbb{E}[C|\mathcal{M}^a], \mathbb{E}[C|\mathcal{M}^{a'}])$ is the discount factor, $\mathbb{E}[C|\mathcal{M}^a]$ and $\mathbb{E}[C|\mathcal{M}^{a'}]$ are the expected cost of making a decision in decision-making system $\mathcal{M}$ modified by $a$ and $a'$, respectively, $\tau: (0, \infty)\times(0, \infty)\to(0, 1]$ computes the discount factor based on the expected cost of making a decision.
\label{def:discounted_DB}
\end{definition}

\subsection{Human-in-the-Loop Decision-Making Systems}

\begin{figure}
    \centering
    \includegraphics[width=0.7\linewidth]{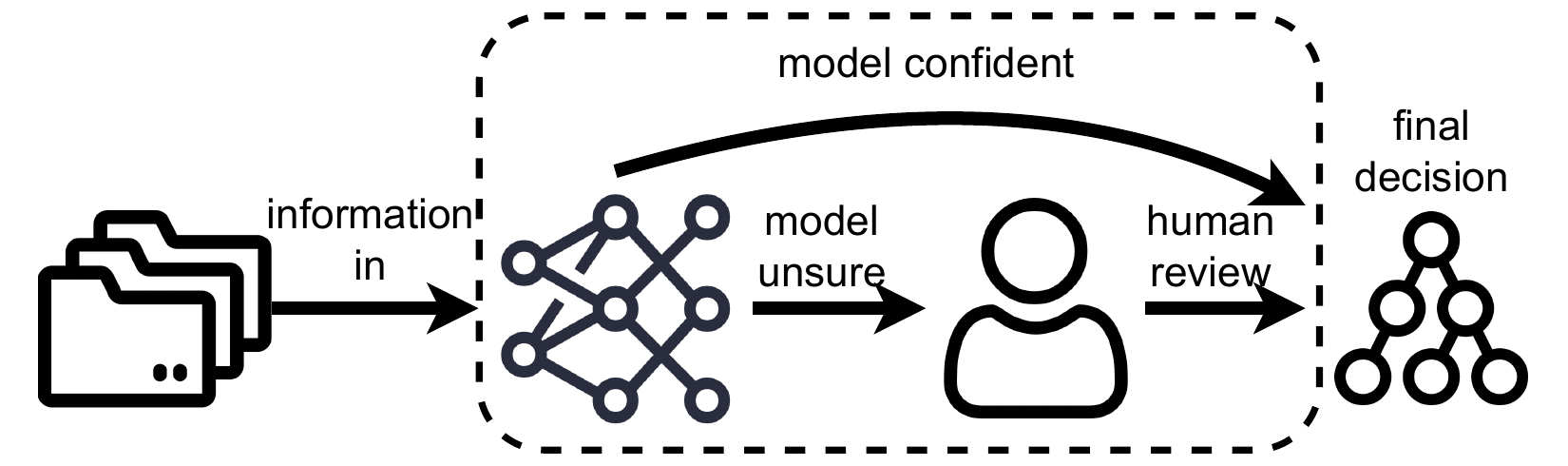}
    \caption{Human-in-the-loop (HITL) decision-making system.}
    \label{fig:HITL}
\end{figure}

The human-in-the-loop decision-making system, illustrated by Figure \ref{fig:HITL}, can be understood through a causal perspective, where both an AI model and human intervention shape the final decision. 
The process starts with an input variable, $X$, representing the data fed into the system. 
This data influences the AI model's output, $M$, which includes a confidence measure. When the model is confident, its decision flows directly to the final decision output, $Y$, bypassing human review. 
However, if the model is uncertain, it passes the decision to a human reviewer, represented by $H$, who considers both the initial data $X$ and the model’s output $M$ before making a judgment. 
The final decision, $Y$, is thus causally determined by either the AI model alone (if confident) or by the combined influence of the model and human input (if the model is unsure).

In this framework, two main causal pathways affect the final decision: a direct path where the model’s confident output directly influences $Y$, and a human-intervention path where the human review shapes $Y$ when the model is uncertain. 
In a decision system $\mathcal{M} = \langle \mathcal{I}, \mathcal{J}, \mathcal{X}, \mathcal{E}, f, \mathbb{P}(\mathcal{E})\rangle$. The decision-making policy $f$ can be modified by actions, which include deploying human-in-the-loop policy, denoted as $a$, and deploying human-only policy, denoted as $a'$. 
Deploying human-in-the-loop policy can increase the efficiency of the decision-making system, at the price of reducing performance. 
The undesired outcome in such system can be making a wrong decision, denoted as $\phi$.
Using Definition \ref{def:discounted_DB}, the degree of blameworthiness can be measure by
\begin{equation}
    DB(a, a') = \gamma(a, a')\delta(a, a'),
    \label{eq:DB_HITL}
\end{equation}
where $\delta(a, a') = \text{max}\left\{0,\ \mathbb{P}(\phi = 1|\mathcal{M}^a) - \mathbb{P}(\phi=1|\mathcal{M}^{a'}) \right\}$. Here, $\delta(a, a')$ measures the extent of deploying HITL policy $f^a$ will increase the probability of making wrong decisions, compared with using human-only policy $f^{a'}$.

\subsection{Attribution of Responsibility for a Specific Undesired Outcome}

Definition \ref{eq:DB_HITL} provides a measure of the overall degree of blameworthiness for an undesired outcome. However, we also aim to attribute responsibility for specific undesired outcomes. For a given outcome $\phi_0$, responsibility is attributed by assessing whether it could have been avoided if an agent had taken an alternative action. In a human-AI collaboration setting, this analysis is not based on real-world probability measures but on the epistemic level the AI agent is expected to maintain to align with responsible AI standards, represented through an adjusted probability measure. In human-only decision systems, responsibility attribution is straightforward as only one agent is involved. Thus, our focus here is on analysing decision-making systems under the HITL policy, where responsibility attribution becomes more complex.

\begin{definition}[Inevitable and avoidable outcomes] 
    Consider a decision-making system represented by the SCM $\mathcal{M} = \langle \mathcal{I}, \mathcal{J}, \mathcal{X}, \mathcal{E}, f, \mathbb{P}(\mathcal{E}) \rangle$, where the decision-making policy is modified by action $a$ under the HITL policy and by action $a'$ under the human-only policy. For a specific undesirable outcome $\phi_0=1$, it can be categorized as follows: 
    \begin{itemize} 
        \item \textbf{Inevitable outcomes}: The event $\phi_0=1$ also occurs in the human-only decision-making system $\mathcal{M}^{a'}$. 
        \item \textbf{Avoidable outcomes}: The event $\phi_0=1$ would not occur in the human-only decision-making system $\mathcal{M}^{a'}$. 
    \end{itemize}
    Furthermore, in the HITL decision-making system, the inevitable outcomes can be categorized as
    \begin{itemize}
        \item \textbf{Flagged outcomes}: event $\psi_0 = 1$, and
        \item \textbf{Unflagged outcomes}: event $\psi_0 = 0$,  
    \end{itemize}
    where $\phi_0$ is $1$ when the AI has requested intervention from human and $0$ if not. 
    \label{def:iao}
\end{definition}

In this definition, inevitable outcomes are those beyond the decision-making capabilities of both human and AI agents, while avoidable outcomes are undesired outcomes that could have been prevented had the AI requested human intervention. Inevitable outcomes can be further categorized as flagged or unflagged, a distinction that may initially seem redundant, as counterfactual reasoning implies the outcome remains unchanged regardless of whether the AI flagged the case. However, this classification is crucial for accurately identifying responsible parties and attributing accountability, aligning with responsible AI standards.

\begin{definition}[Responsibility attribution for different types of outcomes] 
    Consider a decision-\\making system represented by SCM $\mathcal{M} = \langle \mathcal{I}, \mathcal{J}, \mathcal{X}, \mathcal{E}, f, \mathbb{P}(\mathcal{E}) \rangle$, its decision-making policy is modified by action $a$ under the HITL policy, and is modified by action $a'$ under the human-only policy. For a specific undesired outcome $\phi_0=1$, responsibility is attributed as follows:
    \begin{itemize} 
        \item \textbf{Inevitable outcomes:} 
            \begin{itemize} 
                \item \textbf{Flagged}: Responsibility lies with the human who made the decision. 
                \item \textbf{Unflagged}: Responsibility is attributed to both the AI making the decision and the party responsible for designing the flagging mechanism. 
            \end{itemize} 
        \item \textbf{Avoidable outcomes}: Responsibility is assigned to both the AI that made the decision and the party who designed the flagging mechanism.
    \end{itemize}
    \label{def:iao}
\end{definition}

For avoidable outcomes, responsibility is shared between the AI that made the decision and the party who designed the flagging mechanism, as these outcomes could have been prevented if the AI had flagged the case or if the flagging mechanism had been more effective. For inevitable outcomes, the human ultimately bears responsibility, as the system's limitations are constrained by human capabilities. However, it may seem counterintuitive for the AI to also bear responsibility for unflagged inevitable outcomes, since the act of flagging by the AI has no causal effect on the final outcome. This responsibility is not measured by the real-world probability measure $\mathbb{P(\mathcal{E})}$, but rather by a probability measure aligned with responsible AI standards, which assumes that the AI should always consider the potential benefit of requesting human intervention to reduce the likelihood of undesirable outcomes.

\section{Case Study 1: Essay Grading with Large Language Models}

\subsection{Background}

The recent advancements in Large Language Models (LLMs) have enabled their deployment in a variety of educational applications, including automated essay grading \citep{mizumoto2023exploring}. LLMs trained on extensive text corpora demonstrate a remarkable ability to evaluate written content, often producing assessments comparable to those of human graders. However, challenges arise when these models encounter atypical or nuanced writing styles, cultural references, or specific linguistic constructs outside of their training distribution, which may lead to inaccurate grading or biased evaluations \citep{tao2024cultural}.

Integrating LLMs into essay grading systems offers significant potential for efficiency, especially in handling large volumes of student essays. Nevertheless, such integration risks introducing variability in grading quality, particularly in cases where LLMs lack sufficient understanding of complex or context-dependent language use. In this case study, we demonstrate how to assess the degree of blameworthiness using the framework proposed in Section \ref{sec:framework}.

\subsection{Implementation}

In this case study, we evaluate a human-AI collaboration system for essay grading, in which human and AI agents work together to score student submissions. We utilize the ASAP-AES dataset \citep{asap-aes}, specifically focusing on Set-1, which includes persuasive, narrative, or expository essays written by 8th-grade students on the societal impacts of computer use. In response to a prompt, students write a persuasive letter to a local newspaper, arguing whether computers benefit or harm society, with the goal of convincing readers. Essays are graded on a rubric from 1 to 6, based on criteria such as content development, organization, detail, fluency, and audience awareness. Each essay receives two independent scores, and when these scores are close (adjacent), they are summed to determine the final score. For significantly different scores (non-adjacent), an expert adjudicator provides a resolved score.

\begin{figure}[h]
    \centering
    \includegraphics[width=0.45\linewidth]{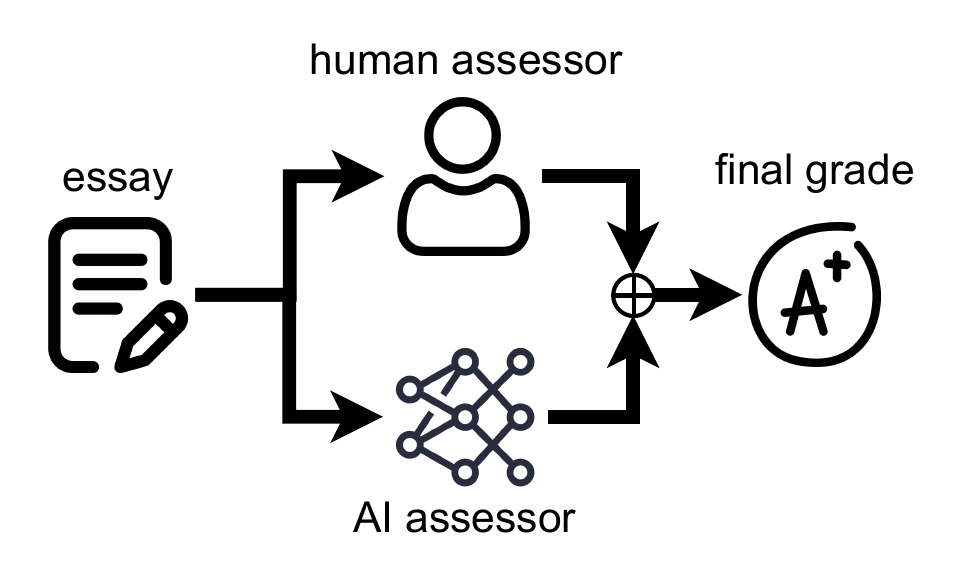}
    \caption{Human-AI collaboration system for essay grading.}
    \label{fig:case_study_1}
\end{figure}

To test the human-AI collaboration framework, we replaced one human assessor with GPT-4 \citep{achiam2023gpt}. The model was provided with the grading rubric and prompted to assign a score based on the essay content. The final grade was calculated by summing GPT-4’s score with that of the human assessor, as illustrated in Figure \ref{fig:case_study_1}.

\subsection{Results}

Using the framework from Section \ref{sec:framework}, we assess the degree of blameworthiness in the human-AI collaborative grading system by comparing the AI-enhanced scores with ground truth scores through the quadratic weighted kappa (QWK) metric \citep{cohen1968weighted}. QWK is a widely used measure of agreement between raters in educational assessment, particularly suited for ordinal scoring tasks like essay grading. This metric penalizes larger discrepancies between scores more heavily, providing a nuanced view of rating consistency. QWK values range from -1 (indicating complete disagreement) to 1 (perfect agreement), with 0 implying no agreement beyond chance. The human-AI grading system achieved a QWK of 0.478, yielding an overall degree of blameworthiness of $1 - 0.478 = 0.522$.

This analysis highlights that while integrating LLMs in human-AI grading systems can improve efficiency, it also risks introducing grading inconsistencies. By systematically attributing responsibility, our framework proves valuable in assessing these systems' reliability and accountability.

\section{Case Study 2: Human-in-the-Loop Pneumonia Detection Using Chest X-ray Images}

\subsection{Background}

Recent advances in AI have significantly enhanced automated medical diagnosis systems, achieving performance levels increasingly comparable to human clinicians \citep{zhou2023foundation, komorowski2018artificial}. However, these models struggle with out-of-distribution cases, where limitations in generalization can lead to diagnostic errors, especially on samples outside the model's training data.

Integrating AI into medical diagnostics offers efficiency gains through rapid processing of large datasets but may introduce new risks, potentially impacting overall system performance. To address these risks, quantifying responsibility for errors between human and AI agents becomes crucial. In this case study, we illustrate how to assess blameworthiness for undesirable outcomes. Using a chest X-ray dataset of 5,863 images labelled as Pneumonia or Normal \citep{kermany2018labeled}, we deploy a human-in-the-loop diagnostic system to detect pneumonia based on chest X-rays

\subsection{Implementation}

In this case study, we examine a human-in-the-loop (HITL) diagnostic system for detecting pneumonia from chest X-ray images. For reproducibility and to simulate different diagnostic capabilities, we use two distinct AI models to represent the roles within the system: a stronger model, ResNet-18 \citep{he2016deep}, mimicking the diagnostic capabilities of a human doctor, and a weaker model, a standard Convolutional Neural Network \citep[CNN;][]{fukushima1980neocognitron}, representing the AI component of the system.

\begin{figure}
    \centering
    \includegraphics[width=0.7\linewidth]{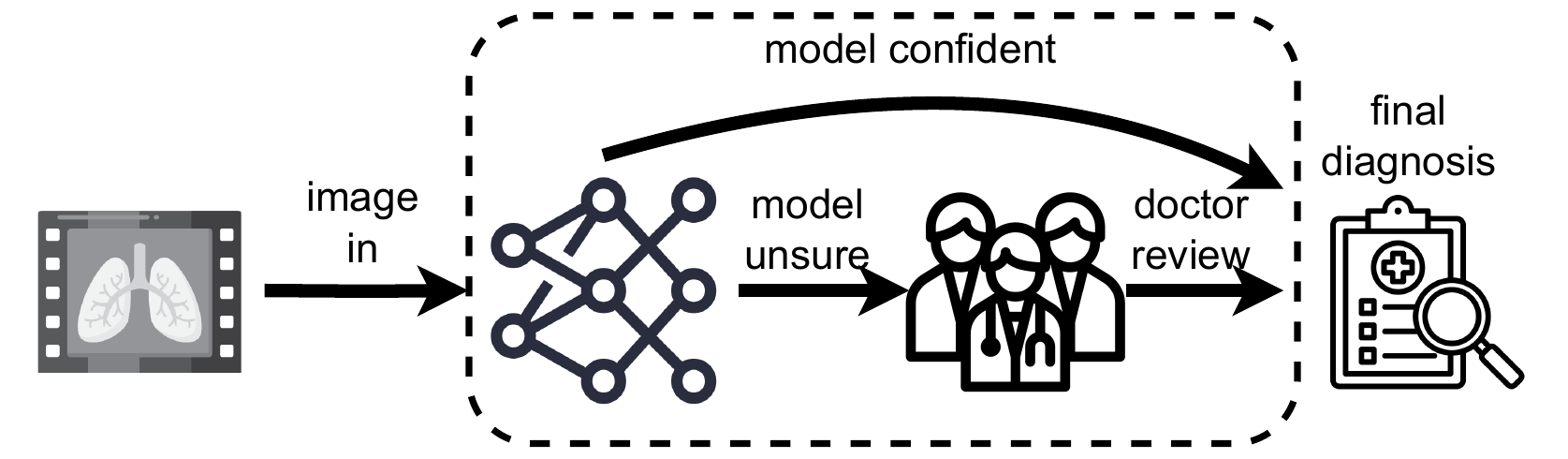}
    \caption{Human-in-the-loop (HITL) diagnostic system for detecting pneumonia from chest X-ray images.}
    \label{fig:case_study_2}
\end{figure}

As shown in Figure \ref{fig:case_study_2}, the HITL system operates as follows: each chest X-ray image is first processed by the weaker AI model, i.e., CNN in this case, which assesses its confidence in making a diagnosis, which is measured by the probability $p$ of detecting pneumonia and pre-defined threshold $l$ and $u$, where $0\le l<u\le1$. If the model is confident in its prediction ($p>u$ or $p<l$), it proceeds to make a decision autonomously. However, if the model detects high uncertainty ($l\le p\le u$), it flags the case and requests human intervention, simulating a collaborative decision-making process. This setup allows us to investigate cases where the AI's autonomy may lead to errors, as well as scenarios where human oversight is involved in the decision.

Using the framework from Section \ref{sec:framework}, we quantify the degree of blameworthiness for the HITL system by comparing the weaker AI model’s decisions with a hypothetical human-only system, represented by ResNet-18.

\subsection{Results}

Applying the framework to the HITL pneumonia diagnostic system, we measure the overall degree of blameworthiness and count for each category of undesired outcomes. The performance of the diagnosis system is measure using F1-score, which is a metric in classification tasks to evaluate a model's accuracy, especially in cases where the data is imbalanced. It is defined as the harmonic mean of precision and recall:

\begin{equation}
    \text{F1-score} = 2 \cdot \frac{\text{Precision} \cdot \text{Recall}}{\text{Precision} + \text{Recall}},
\end{equation}
where 
\begin{equation}
    \text{Precision} = \frac{\text{TP}}{\text{TP} + \text{FP}}, \quad \text{Recall} = \frac{\text{TP}}{\text{TP} + \text{FN}}.
\end{equation}

Following the deployment of a human-in-the-loop (HITL) policy, the diagnostic system’s F1-score decreased from 0.896 to 0.831, with this drop indicating the degree of blameworthiness associated with implementing the HITL system.

For inevitable outcomes, errors persist despite human oversight, revealing cases where both human and AI agents struggle with challenging, out-of-distribution data. Specifically, 2 inevitable errors were flagged by the AI for human intervention, while 83 inevitable errors were not flagged, suggesting overconfidence in the AI’s predictions. A well-calibrated AI model and an improved flagging mechanism could reduce responsibility for inevitable cases that are not flagged.

Among the 73 avoidable errors, responsibility falls primarily on the AI and those responsible for designing the flagging mechanism, as these errors could have been prevented if the AI had flagged them for human intervention. Here, the human is not blameworthy, since these errors were due to the AI’s failure to request intervention.

Overall, this analysis reveals that, while the HITL approach can improve efficiency, it introduces risks when the AI is overconfident in its predictions and the flagging mechanism lacks precision. By precisely defining responsibility for specific outcomes, this study demonstrates the utility of our framework in evaluating accountability for both AI and human agents within HITL diagnostic systems. These findings underscore the need for further refinements in model calibration and flagging mechanisms to optimize human-AI collaboration in high-stakes applications.

\section{Conclusion}

This paper presented a structured causal responsibility attribution framework designed for human-AI collaboration decision-making systems, where AI models and human agents collaborate to make critical decisions. As AI integration continues to expand in sectors such as healthcare and education, the need for precise responsibility attribution becomes increasingly important, particularly for undesired outcomes that can impact trust, accountability, and ethical AI deployment.

The formalization of outcomes and blameworthiness in our framework provides a rigorous approach to assess the impact of an agent's actions on undesired outcomes, enabling a systematic evaluation of responsibility. By defining an outcome as a measurable function within an SCM, we quantify blameworthiness by comparing the probabilities of outcomes under different actions. We also introduced a discounted blameworthiness measure, which adjusts for the efficiency improvements that an action might bring, reflecting a balanced view of accountability where both performance and cost are considered.

Our framework distinguishes between inevitable and avoidable outcomes, with avoidable outcomes further divided into flagged and unflagged categories. This classification enables precise responsibility attribution: inevitable outcomes are attributed to limitations inherent to both AI and human agents, while avoidable outcomes specify responsibility based on whether human intervention was flagged or bypassed. Specifically, responsibility for avoidable errors may lie with the AI, the human agent, or the designers of the flagging mechanism, depending on whether intervention was requested. This approach addresses key issues in existing methods, such as Shapley values, which tend to disproportionately blame agents contributing more to an outcome and rely on real-world measures of blameworthiness that may not align with responsible AI standards.

Using this approach, we demonstrated the framework’s applicability through two case studies: ASAP-AES essay grading and pneumonia detection from chest X-ray images. These examples illustrate how our framework systematically assesses causal responsibility for each agent within HITL systems, offering insights to inform future improvements in AI design and human-AI collaboration strategies. However, this study does not address measuring blameworthiness for sequential decisions that align with responsible AI standards, which remains a direction for future research.

Our causal responsibility attribution framework represents a step forward in accountability for human-AI collaboration systems, promoting transparent and ethical AI deployment in high-stakes settings. By integrating causal reasoning with aligned with responsible AI standards, this work paves the way for responsible human-AI collaboration, laying a foundation for trust in increasingly autonomous systems.

\bibliography{ref}

\end{document}